\documentclass[sigconf, nonacm, 10pt]{arxiv_acmart}
\AtBeginDocument{%
  \providecommand\BibTeX{{%
    Bib\TeX}}}

\usepackage{balance}

\usepackage{xcolor}
\def\BibTeX{{\rm B\kern-.05em{\sc i\kern-.025em b}\kern-.08em
    T\kern-.1667em\lower.7ex\hbox{E}\kern-.125emX}}
\usepackage[english]{babel}
\usepackage{graphicx}
\usepackage{amsmath,amsfonts} %
\usepackage{textcomp}
\usepackage{xcolor}
\usepackage{bbold}
\usepackage{multicol}
\usepackage{multirow}
\usepackage[T1]{fontenc}
\usepackage{listings}
\usepackage[utf8]{inputenc}
\usepackage{bm}
\usepackage{gensymb}
\usepackage{tikz}
\usepackage{xspace}
\usepackage{soul}
\usepackage{ragged2e}
\usepackage{enumitem}
\usepackage{microtype}
\usepackage{booktabs}
\usepackage{subcaption}
\usepackage{xfrac}
\usepackage{float}
\usepackage{lipsum}
\usepackage{balance}
\usepackage{algorithm}
\usepackage{algorithmic}
\usepackage[algo2e,ruled,vlined,noend,linesnumbered]{algorithm2e}
\usepackage{array}
\usepackage{balance}

\newcommand{\latinphrase}[1]{\textit{#1}}

\newcommand{\ie}{\latinphrase{i.e.,}\xspace}
\newcommand{\eg}{\latinphrase{e.g.,}\xspace}

\DeclareMathOperator*{\argmax}{argmax}
\DeclareMathOperator*{\argmin}{argmin}

\newcommand{\MAC}{\texttt{MAC}\xspace}
\newcommand{\Bandwidth}{\texttt{Bandwidth}\xspace}
\newcommand{\Latency}{\texttt{Latency}\xspace}
\newcommand{\Energy}{\texttt{Energy}\xspace}

\newcommand{\oursystem}{\textsl{Centaur}\xspace}
\newcommand{\iot}{UCD\xspace}
\newcommand{\iots}{UCDs\xspace}
\newcommand{\ap}{AP\xspace}
\newcommand{\aps}{APs\xspace}

\newcommand{\revs}[1]{\textcolor{black}{#1}}

\DeclareRobustCommand\blackline
{\tikz[baseline=0.0ex]\draw[black, line width=0.8pt] (0,0.1) -- (0.23,0.1);}

\makeatletter
\newcommand{\algorithmfootnote}[2][\footnotesize]{%
  \let\old@algocf@finish\@algocf@finish
  \def\@algocf@finish{\old@algocf@finish
    \leavevmode\rlap{\begin{minipage}{\linewidth}
    #1#2
    \end{minipage}}%
  }%
}
\makeatother

\setcopyright{none}
\renewcommand\footnotetextcopyrightpermission[1]{} 
\newcommand\blfootnote[1]{%
  \begingroup
  \renewcommand\thefootnote{}\footnote{#1}%
  \addtocounter{footnote}{-1}%
  \endgroup
}
\begin{document}

\title{Enhancing Efficiency in Multidevice Federated Learning through Data Selection}

\author{Fan Mo}
\email{mofan1992@gmail.com}
\affiliation{\institution{Imperial College London, }\country{London, UK \\ {\small (Work done at Nokia Bell Labs)}}}
\author{Mohammad Malekzadeh}
\email{{mohammad.malekzadeh@nokia-bell-labs.com}}
\affiliation{\institution{Nokia Bell Labs}\country{Cambridge, UK}}
\author{Soumyajit Chatterjee}
\email{soumyajit.chatterjee@nokia-bell-labs.com}
\affiliation{\institution{Nokia Bell Labs}\country{Cambridge, UK}}
\author{ Fahim Kawsar}
\email{fahim.kawsar@glasgow.ac.uk}
\affiliation{\institution{\mbox{University of Glasgow}}\country{Scotland, UK}}
\author{Akhil Mathur}
\email{akhilmathurs@gmail.com}
\affiliation{\institution{\mbox{Nokia Bell Labs}}\country{Cambridge, UK}}

\renewcommand{\shortauthors}{Fan Mo, Mohammad Malekzadeh, Soumyajit Chatterjee, Fahim Kawsar, and Akhil Mathur}
\begin{abstract}
Ubiquitous\blfootnote{\normalsize To be presented in the 10$^{th}$ ACM/IEEE Symposium on Edge Computing (SEC'25), December 2025, Arlington, VA, USA.} wearable and mobile devices provide access to a diverse set of data. However, the mobility demand for our devices naturally imposes constraints on their computational and communication capabilities. A solution is to locally learn knowledge from data captured by  ubiquitous devices, rather than to store and transmit the data in its original form. In this paper, we develop a federated learning framework, called \oursystem, to incorporate {\em on-device data selection} at the edge, which allows {\em partition-based training} of a deep neural nets through collaboration between constrained and resourceful devices within the {\em multidevice ecosystem of the same user}. We benchmark on five neural net architecture and six datasets that include image data and wearable sensor time series. On average, \oursystem achieves $\sim$19\% higher {\em classification accuracy} and $\sim$58\% lower {\em federated training latency}, compared to the baseline. We also evaluate \oursystem when dealing with {imbalanced non-iid data}, {client participation heterogeneity}, and different {mobility patterns}. To encourage further research in this area, we release our code at \textcolor{blue}
 {\href{https://github.com/Nokia-Bell-Labs/data-centric-federated-learning}
 {github.com/nokia-bell-labs/data-centric-federated-learning}}.
\end{abstract}





\maketitle
\section{Introduction}\label{sec_intro}

\begin{figure}[]
    \centering
    \begin{subfigure}[t]{\columnwidth}
    \centering
        \includegraphics[width=\columnwidth]{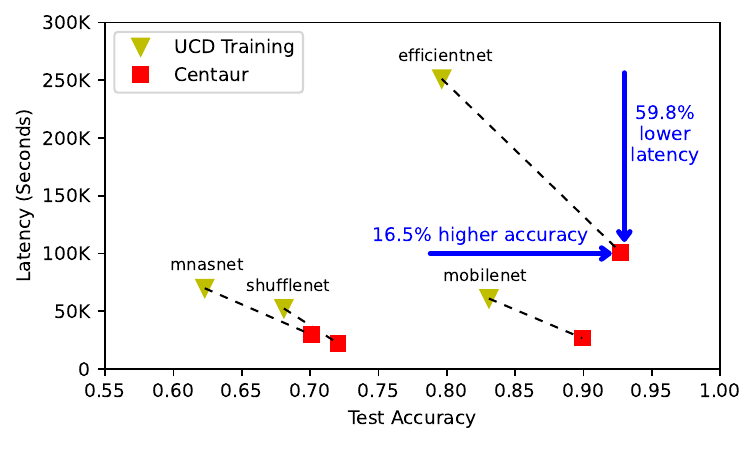}
        \label{fig:intro_sum_a}
    \end{subfigure}%
    
    \begin{subfigure}[t]{\columnwidth}
    \centering
        \includegraphics[width=\columnwidth]{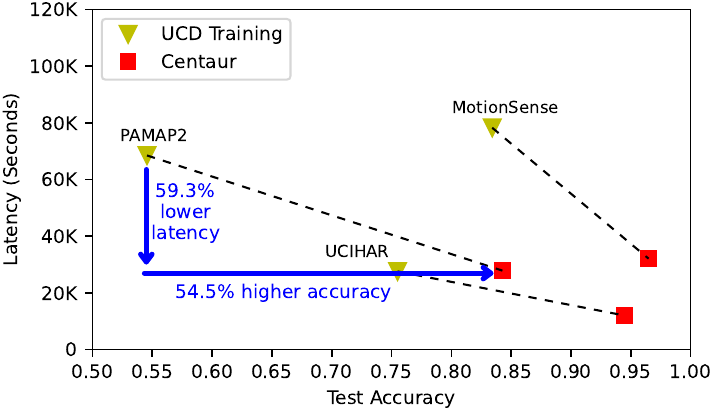}
        \label{fig:intro_sum_b}
    \end{subfigure}
    \caption{The latency of federated training versus classification accuracy on test dataset. We compare \oursystem with standard federated training that only trains the classifier on ubiquitous constrained devices~(\iots) without data selection and partition-based training. (Top plot) Four models were tested on image data from CIFAR-10, and (bottom plot) a model was tested on three datasets of wearable sensor time-series.}
    \label{fig:intro_sum}
\end{figure}

With the growing trend of utilizing ubiquitous devices in personal and industrial environments, federated learning~(FL) enables the discovery of new patterns in personal data by training deep neural networks~(DNNs) on data captured by various users in a more private manner~\cite{floriginal, flsurvey}. Especially, in a multidevice ecosystem, the owner can seamlessly share data across the devices within their trusted environment while protecting data privacy~\cite{cho2022flame}. This allows us to utilize the capabilities of multiple devices to participate in the FL process. However, wearable and ubiquitous devices have limited computing resources, little on-device storage, and inconsistent network connectivity. 

Most of the existing FL frameworks do not account for constraints of personal devices and instead assume either modern smartphones~\cite{hard2018federated, niu2020billion}, or edge devices with DNN accelerators~\cite{cho2022flame, zhang2022federated}, as target client devices. Although smartphones and accelerators are less powerful than cloud machines, they are nevertheless equipped with mobile GPUs, possess gigabytes of runtime memory, and have fairly stable connectivity to a central server, which simplifies the requirements of DNN training~\cite{yang2021flop, lin2020ensemble}.

In this paper, we propose a novel FL framework to include {\em ubiquitous constrained devices} (\iots), \ie wearable devices such as earbuds, glasses, or rings, or distributed edge devices such as environmental cameras or industrial sensors. We believe \iots are becoming the primary {\em data-producing} devices for both individuals and industries, due to various on-device sensors, such as cameras or inertial measurement units. These devices offer a wealth of spatiotemporal data that is frequently absent when focusing solely on smartphones or plugged-in devices. For example, people often opt for wristbands, smart rings, or earbuds over carrying a smartphone in their hand or pocket while engaging in outdoor, industrial, or sports activities~\cite{seneviratne2017survey, mo2021adapting}, and sensors distributed across farms or forests remotely monitor crops and livestock to detect hazards~\cite{papst2021exploring}. The main {\em motivation} of our work is the fact that the abundance and diversity in data captured by ubiquitous devices can facilitate learning more effective models for emerging applications in personal and industrial environments. 

There are several significant {\em challenges}. Usually, distributed or wearable \iots use an {\em access point}~(\ap), like a smartphone or a router, as a relay to connect to a {\em central server} in the cloud. Data captured by a \iot is communicated to the \ap using communication protocols like Bluetooth Low Energy~(BLE). In outdoor settings under mobility, \iots may not always be connected to the companion \ap, thus inaccessible to the central server. Beside this, the constrained data storage and computing resources will restrict the size and functionality of DNNs that can be trained locally on \iots. A naive solution is to transfer all data from \iots to their respective \aps and perform training of DNNs on \aps. Yet, storing {\em all} data could exceed the device's memory, especially when \iots disconnect from their \ap. To demonstrate these points, in \S\ref{sec_motiviating_rpi}, we conduct a motivating study on a real testbed to identify the essential design requirements that can enable the deployment of FL on resource-constrained devices. To tackle these challenges effectively, we believe FL solutions relying on on-device training should dynamically distribute data and computations between \iots and their companion \aps.

\begin{table*}[t!]
\centering
\caption{Comparison with related work. \textit{(1) Model Partitioning} splits the model into an encoder and a classifier. \textit{(2) Data Selection} trains on a subset of available data. \textit{(3) On-device Training} considers backpropagation on constrained devices. \textit{(4) Spatiotemporal Coverage} includes mobility patterns of portable devices.}\label{tab_related_work}
\resizebox{\textwidth}{!}{%
\begin{tabular}{lcccc}
& {Model Partitioning}  & {Data Selection} & {On-device Training} & {Spatiotemporal Coverage}  \\
{Multitier FL \cite{abad2020hierarchical, liu2020client, abdellatif2022communication, cho2022flame, gupta2018distributed}}& \multirow{1}{*}{\texttimes} & \multirow{1}{*}{\texttimes} & \multirow{1}{*}{\texttimes} & \multirow{1}{*}{\texttimes} \\ \cline{1-5}
{Split FL \cite{diao2020heterofl, zhang2021fedzkt, zhang2021parameterized, li2021hermes, dai2022dispfl, bibikar2022federated}}&   \multirow{1}{*}{\checkmark} & \multirow{1}{*}{\texttimes}    &  \multirow{1}{*}{\texttimes}                                &  \multirow{1}{*}{\texttimes}               \\ \cline{1-5}
{On-device Training \cite{lee2019intermittent, islam2019zygarde, fedorov2019sparse, banbury2021micronets, xu2022etinynet, liberis2021munas, lin2022device, kopparapu2022tinyfedtl}}& \multirow{1}{*}{\texttimes}                        &       \multirow{1}{*}{\texttimes} &            \multirow{1}{*}{ \checkmark  }      &  \multirow{1}{*}{ \checkmark  }                         \\  \cline{1-5}
{\textbf{\oursystem} (ours)}     & \checkmark                            & \checkmark             & \checkmark                       & \checkmark                           \\
\end{tabular}
}
\end{table*}

In \S\ref{sec_metod} and  Figure~\ref{fig:PbFL}, we introduce {\bf \oursystem}: a federated learning framework that orchestrates local training among \iots and \aps by integrating {\bf on-device data selection} with {\bf partition-based model training}. 

{\bf (A)}~Considering computation constraints, we initialize the DNN in two partitions: an {\em encoder}~(\ie~feature extractor) to be only trained on \aps, followed by a lightweight {\em classifier} to be trained on both \aps and \iots~(\S\ref{subsec:model_init}). This allows \oursystem to utilize all the data available on \iots for fine-tuning the DNN while occasionally storing and transmitting a portion of the data to the \aps to re-train the entire DNN and adjust to the runtime variability in data distributions~\cite{gupta2018distributed, thapa2022splitfed,weiss2016survey}. 

{\bf (B)}~Considering memory and connectivity constraints, we perform data selection~\cite{jiang2019accelerating} by analyzing the {\em training loss} and the {\em gradients norm} of the classifier part, to decide which data points captured by \iots contribute more to the training of which part of the DNN. Through data selection, data points are categorized as either of (i) {\em discarded} if they have very low loss values, (ii) {\em kept} locally on the \iot to train the classifier part if their loss values or gradients norm are not very high, and (iii) {\em transmitted} to the \ap to train both encoder and classifier part if they cause high values for both loss and gradients norm~(\S\ref{subsec:data_selection}). We assume that each pair of \iot and \ap belongs to the same client, thus transmitting data from a \iot to its \ap does not violate the clients' privacy.


{\bf (C)} We benchmark on five DNNs architectures and six datasets of two different modalities, and compared to existing FL alternatives. Our results show that \oursystem saves more {\em bandwidth} by reducing the communication cost associated with offloading samples from a \iot to the \ap, and also reduces the {\em latency} of training (and accordingly the {\em energy} consumption) on both \iot and \ap. As a prime example of our experimental results,  \figurename~\ref{fig:intro_sum} shows the training latency versus classification accuracy of performing FL for two different tasks: (1) four benchmark DNNs, with their encoder pre-trained on ImageNet~\cite{deng2009imagenet}, and then trained for CIFAR10~\cite{krizhevsky2009learning} image classification with 100 FL clients, and (2) a benchmark ConvNet~\cite{chang2020systematic}, trained from scratch, on three human-activity recognition datasets with 20 FL clients. In both figures, we compare \oursystem with the baseline of standard FL on \iots without our implemented strategies. \oursystem, through data selection and partition-based model training, achieves up to $19\%$ higher accuracy and $58\%$ lower latency, on average. 

The {\bf key contribution} of our work is a better integration of ubiquitous constrained devices into federated learning by leveraging the advantages of multi-device ecosystems. We achieve this by introducing a customizable data selection scheme and a partition-based training approach, which collectively reduce computational and communication costs while enhancing model accuracy. Our experiments demonstrate that \oursystem effectively minimizes storage requirements and training time through efficient data selection. Furthermore, our analysis accounts for the real-world spatiotemporal mobility of edge devices in FL.  Empirical evaluations, covering cost, data imbalance, participation heterogeneity, and connection probability, show that \oursystem consistently delivers higher efficiency, achieving improved accuracy at lower costs across diverse scenarios.

\section{Related Work}\label{sec_related}
Table~\ref{tab_related_work} compares \oursystem with major prior work on different aspects of multidevice FL. Compared to the standard edge--server FL, edge--\textit{access point}--server FL~\cite{liu2020client, abad2020hierarchical} considers a middle layer, \eg a cellular base station, to orchestrate FL training across clients and a central server. Clients can participate with several devices~\cite{cho2022flame, diao2020heterofl, zhang2021fedzkt}, which achieve better trade-offs between accuracy and bandwidth consumption. However, these works assume the local training of the entire model on edge devices, without any resource limitations. They also do not utilize the potential of resource-rich devices to assist a resource-constrained device during training. They use \aps only for aggregation and communication, which makes FL impractical in scenarios involving \iots.

Other works consider inference only (\ie~forward pass)~\cite{fedorov2019sparse, banbury2021micronets, xu2022etinynet, liberis2021munas}, sparse training~\cite{li2021hermes, bibikar2022federated}, or heterogenous architectures across clients of different capabilities~\cite{kang2023nefl, park2024federated}. Sparse training or pruning requires computational resources~(especially memory) that \iots do not have. Few works propose solutions for full on-device training (\ie~forward and backward pass)~\cite{lin2022device}, but not for FL scenarios as they often make impractical assumptions like the availability of large datasets for network architecture search~\cite{liberis2021munas}, or unconstrained memory for training the full model~\cite{diao2020heterofl, zhang2021fedzkt}. Transfer learning can be used on \iots to fix the encoder architecture, which in turn restricts the extraction of newer features, and only trains the last layer for the personalized training~\cite{zhang2021fedzkt}.

Without utilizing the power of \aps, training the entire model on \iots requires a large sample size to learn generalized features~\cite{li2021hermes, bibikar2022federated}. Simply considering different sub-models~\cite{diao2020heterofl,kang2023nefl, park2024federated} requires more sophisticated aggregation strategies and shallower models for resource-constrained devices, which may not be capable of learning the complex patterns from the data.
Recent works enable training on micro-controllers with limited memory, such as TensorFlow Lite Micro~\cite{david2021tensorflow} or Tiny Training Engine~\cite{lin2022device}. These works do not consider FL as a potential use case and do not account for the advantages of a multidevice ecosystem.

\section{A Motivating Study} \label{sec_motiviating_rpi}
\begin{figure*}[t!]
    \centering
    \includegraphics[width=\textwidth]{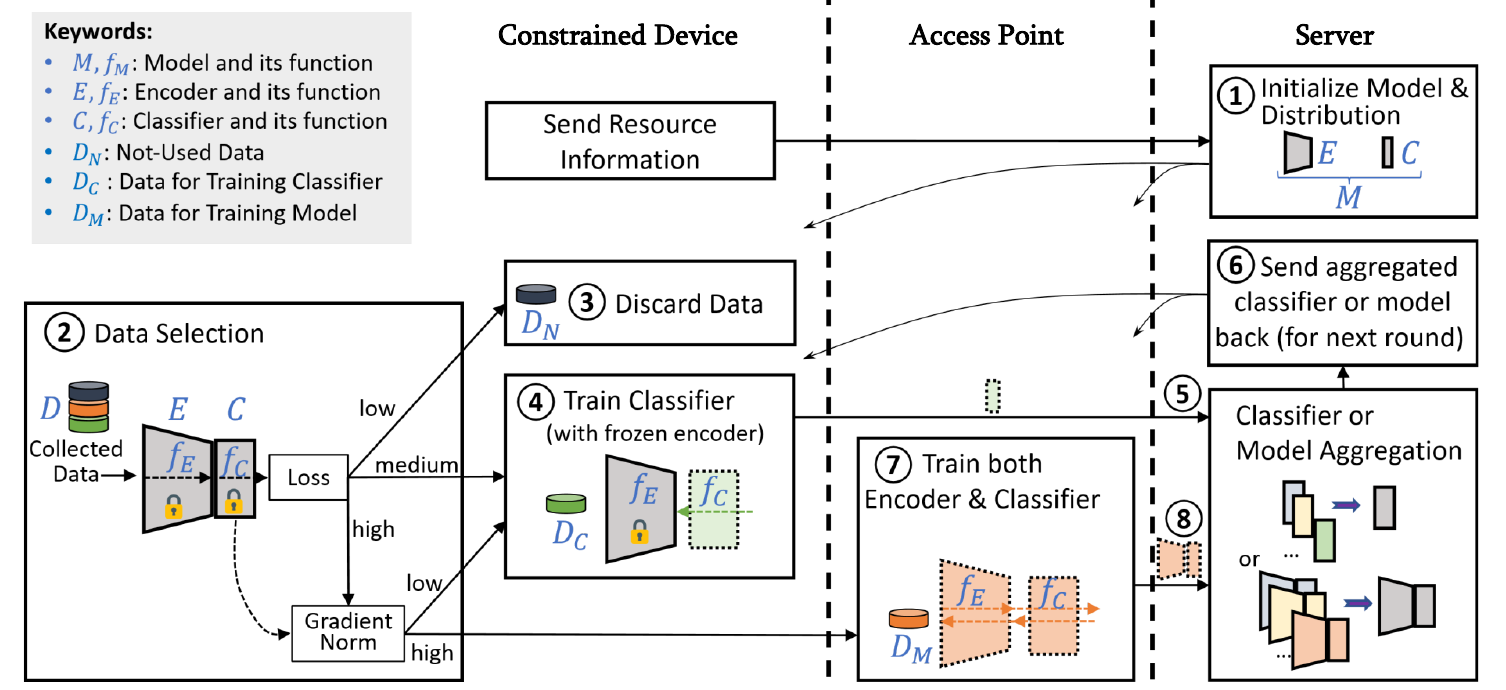}
    \caption{   The overview of \oursystem{}, including model initialization (\S~\ref{subsec:model_init}), data selection (\S~\ref{subsec:data_selection}), and partition-based training and aggregation (\S~\ref{subsec:fl_training}). We explain the set up in \S~\ref{subsec:framework_overview} and elaborate the details of Steps~\textcircled{\raisebox{-.9pt} {1}} to~\textcircled{\raisebox{-.9pt} {8}}.
    }
    \label{fig:PbFL}
\end{figure*}

As depicted in Figure~\ref{fig:PbFL}, a {\em ubiquitous constrained device}~(\iot) is \revs{a small, low-power portable device with restricted hardware and connectivity resources. Examples include smart rings, earbuds or glasses (personal wearables) and tiny environmental or industrial sensors. Each UCD connects to the user’s personal {\em access point}~(AP), usually over a low-power link such as Bluetooth Low Energy.} An AP is 
a resourceful device that has much better hardware and connectivity capabilities than its corresponding \iot (for instance, a smartphone, home router, or gateway). We assume the \ap is always part of the user's device ecosystem that allows the \iot to transmit data to the \ap without violating the user's privacy. \revs{We generally assume one \ap serves multiple \iots (\eg a person’s phone might pair with a smartwatch, earbuds, and other sensors simultaneously). Indeed, modern homes routinely host on the order of 20–30 connected devices, so a single \ap often handles many \iots.}

A {\em server} is a central entity owned by a service provider that orchestrates FL over several clients. Considering a deep neural network architecture, an {\em encoder} ($E$) is the first part of the DNN that extracts the features from input data. A {\em classifier} ($C$) is the second part of the DNN, which performs the final classification using the encoded features. 
We consider supervised learning tasks where $\mathbf{X}$ denote the input data and $\mathbf{y}$ denote its label.

\begin{figure}
	\centering
    \subfloat[The Testbed\label{fig:rpi_setup}]{
                             \includegraphics[
                             width=0.3\columnwidth, keepaspectratio]{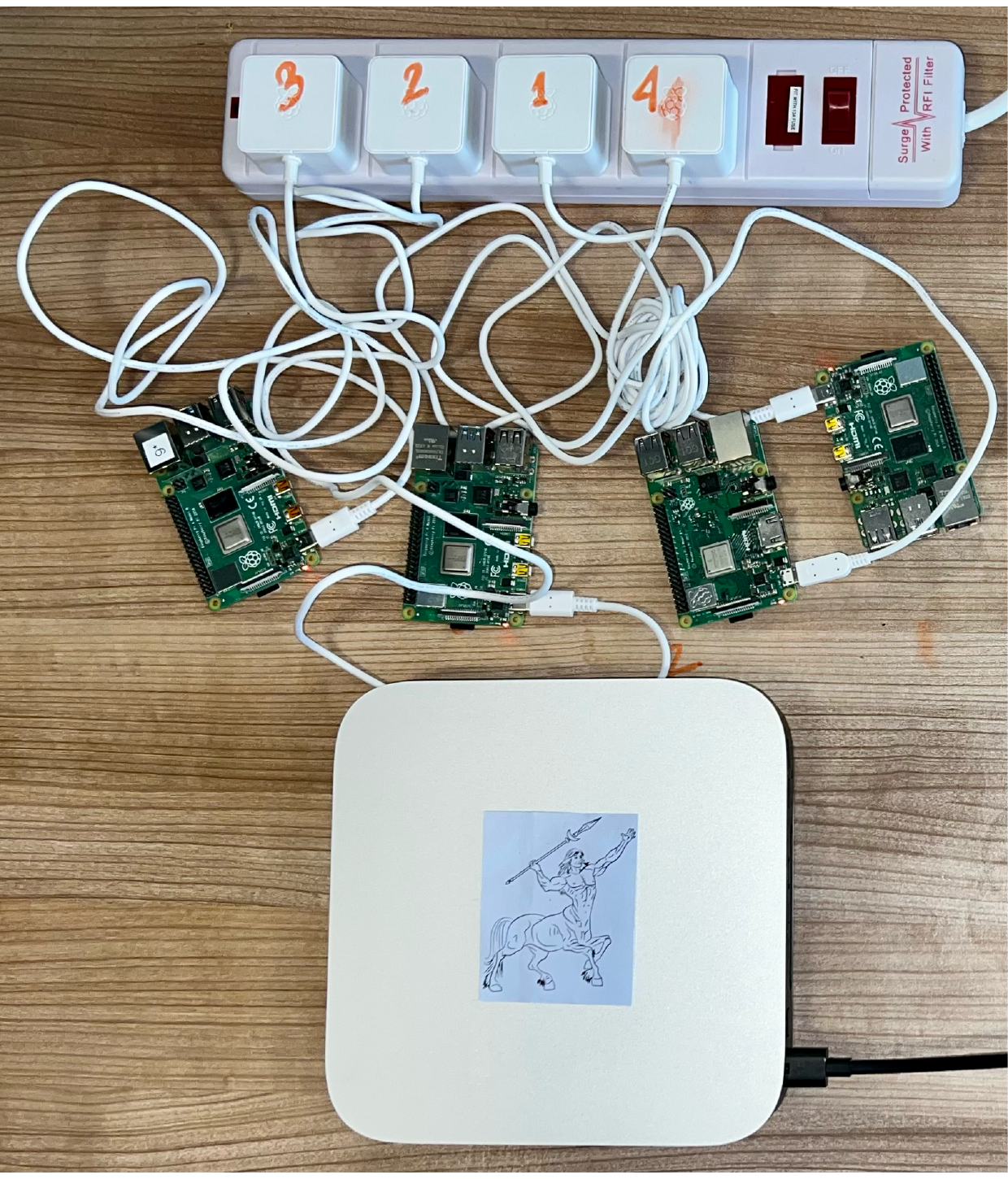}
    }
	\subfloat[Memory Consumption\label{fig:mem_consu}]{
		\includegraphics[
  width=0.68\columnwidth, keepaspectratio]{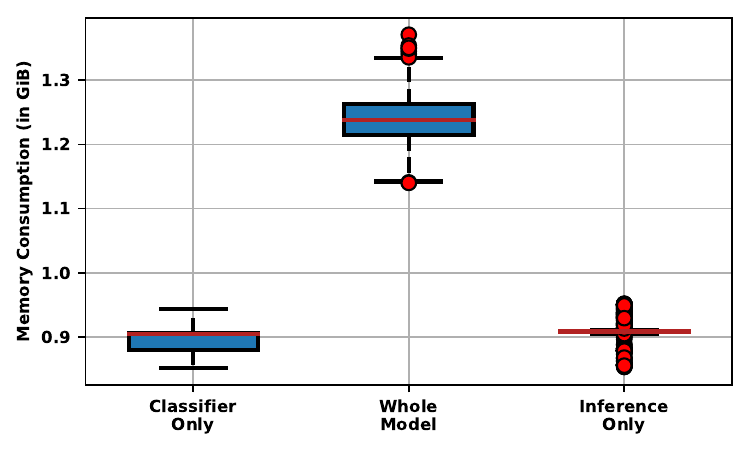}
	}
	\caption{On four RaspberryPis, we measure memory consumption of running FL for when (i) only the classifier is trained, (ii) the entire model is trained, and (iii) the entire model runs inference only.}
	\label{fig:resource_consu}
\end{figure}

{\bf Setup.} To identify the key design choices, we create a testbed with four RaspberryPi 4 Model B as edge devices~\cite{velasco2018performance, drakopoulos2019real}; all running the 64-bit Raspbian OS with total secondary storage of $32$GB. We set up four clients with varying primary memories: C1, C2, C3, and C4  having $1$GB, $2$GB, $4$GB, and $2$GB, respectively. \revs{Indeed, Raspberry Pi devices are more capable than many wearables; however, they serve as a widely adopted, reproducible proxy for constrained clients~\cite{cho2022flame}, enabling controlled CPU/memory benchmarking while we contextualize our findings with real-world wearable hardware specifications and multitasking considerations.} For analyzing the CPU and memory consumption of the clients, we use logs of \texttt{/proc/stat} and \texttt{free} operating-system calls, respectively.  We use MobileNetV3~\cite{howard2019searching}, pre-trained on ImageNet, as the encoder followed by two fully-connected layers with an intermediate drop-out layer. Using CIFAR10, we measure the resources and time needed for both on-device training (per epoch) and inference. We implement FL with Flower~\cite{beutel2022flower} and use a desktop computer as the server, as shown in Figure~\ref{fig:resource_consu}a.

{\bf Key Observations.} We run FedAvg~\cite{floriginal} for $100$ rounds with four RaspberryPi devices, considering two scenarios: (1) when only the classifier part is trained, and (2) when the entire model (encoder and classifier) is trained. When running our experiments for the first scenario, we observed that the standard FL runs successfully over all four clients. However, for the second scenario, client C1 with 1GB of primary memory always failed to participate in the FL process. This is primarily due to the significant rise in the CPU and memory consumption when the entire model (both encoder and classifier) are trained on the device (see \figurename~\ref{fig:resource_consu}).

Notably, 1GB of primary memory is still significant in terms of memory capability, and many \iots may have significantly less memory than this. Therefore, it is clear that in such cases, it is much more challenging to deploy standard FL over \iots. Moreover, training only the classifier consumes far less resources and is even comparable to the resource usage required for a forward pass. We noticed a similar pattern for time consumption, where training the classifier only is $50\times$ time efficient compared to training the entire model. One alternative solution to decrease the overall training time is to reduce the training data size by randomly discarding samples. However, we observe that such an approach comes at a cost of significantly reduced accuracy, which is detrimental to the overall system.

\section{Method} \label{sec_metod}

In Figure~\ref{fig:PbFL}, we introduce \oursystem{}, our federated learning framework. Here, we elaborate on each component in detail.

\label{subsec:framework_overview}
\textbf{Step~\textcircled{\raisebox{-.9pt} {1}}}: The server receives the information about \iots' resource. Accordingly, the server initializes a DNN model $M$ that is formed of an encoder $E$ and a classifier $C$, and then distributes the $E$ and $C$ to the \aps and subsequently to \iots. We explain model initialization details in \S\ref{subsec:model_init}.

\textbf{Step~\textcircled{\raisebox{-.9pt} {2}}}: \iots start FL by performing data selection on the collected data $D$, based on the computed loss value via the forward pass and the magnitude of the gradient of the classifier's last layer. Each sample point will be categorized in one of (i) $D_{N}$ to be discarded, (ii) $D_C$ for locally training $C$ on the \iot, or (iii) $D_M$ to be transmitted to \ap for training $M$. We explain data selection details in~\S\ref{subsec:data_selection}.

\textbf{Step~\textcircled{\raisebox{-.9pt} {3}}}: The \iot discards data $D_N$ in the current epoch without any further training steps or backward pass on them. 

\textbf{Step~\textcircled{\raisebox{-.9pt} {4}}}: The \iot performs backward pass through the classifier $C$ on data $D_C$, while the encoder $E$ is frozen. Then, the \iot updates the parameters of $C$ via computed gradients. Notice that, in practice, the gradient computation done in Step~\textcircled{\raisebox{-.9pt} {2}} can be reused for this update. 

\textbf{Step~\textcircled{\raisebox{-.9pt} {5}}}: \iots share their updated classifier $C$ to the server, and the server performs aggregation on all $C$ from participating \iots to obtain one aggregated classifier.

\textbf{Step~\textcircled{\raisebox{-.9pt} {6}}}: The server sends the aggregated classifier back to the \iots and \aps.

\textbf{Step~\textcircled{\raisebox{-.9pt} {7}}}: With the updated classifier, the \ap trains the encoder $E$ together with the classifier $C$ atop $E$ on data $D_M$. That is, $D_M$ is fed for a forward pass and then a backward pass throughout to update $M$.

\textbf{Step~\textcircled{\raisebox{-.9pt} {8}}}: The updated $M$ (including $E$ and $C$) is shared from the \ap to the server. After collecting updated $M$ from all participating \aps, the server performs model aggregation to obtain the aggregated encoder $E$ and classifier $C$ and distributes them back to \iots and \aps as in Step~\textcircled{\raisebox{-.9pt} {6}}. See \S~\ref{subsec:fl_training} for more details of FL training procedures.

Steps~\textbf{\textcircled{\raisebox{-.9pt} {2}}} to~\textbf{\textcircled{\raisebox{-.9pt} {8}}} will be repeated for $R/2$ rounds to update $C$ on \iots and $E$ on \aps. Algorithms~\ref{alg:training_aggr} shows the process of partition-based training and aggregation in \oursystem. 

\revs{Since full backpropagation involves a forward pass, the storage of all intermediate activations, and a backward pass through every layer to compute gradients, it is the most compute- and memory-intensive phase of model training. \oursystem mitigates this by freezing the deep encoder and training only the lightweight classifier head on \iots. Each device performs a single forward pass through the fixed encoder and backpropagates only through the shallow fully-connected classifier, which has orders of magnitude fewer parameters. This design significantly reduces the on-device workload. Our motivationa study in \S\ref{sec_motiviating_rpi} confirmed this trend, with classifier-only training running several times faster and using far less memory than updating the entire network. Thus, \oursystem enables effective training on local data while keeping \iot training lightweight and efficient.}

\subsection{\textbf{Model Initialization}}
\label{subsec:model_init}

To partition a DNN into an encoder $E$ and a classifier $C$, one option is to use pre-trained benchmark DNNs; considering numerous well-trained models.
Usually, DNN architectures from a candidate space of pre-trained models have already been well developed, with efficient quantization and compression capabilities, to be deployed on \iots for inference~\cite{fedorov2019sparse, banbury2021micronets, xu2022etinynet, liberis2021munas, li2021hermes}. We aim to utilize the full capability of  \iots for on-device training of the classifier $C$. Thus, we need to design the classifier such that it can run on the limited memory and computation power available on these devices. 
To achieve these, we can iteratively look into candidate architectures that fit in the limited memory available on these \iots.
A robust mechanism in this direction is to obtain a history of memory usage of a particular device, which reflects the typical memory availability on the \iot. 

\revs{To make the selection process explicit, we consider a set of lightweight architectures $\{m_k\}$: a small set of widely used mobile convolutional models (\eg SqueezeNet and MobileNet-style variants), alongside fully connected networks with varying depth (1–3 layers) and width (16–128 neurons). These architectures are chosen because they span the typical memory–compute trade-off space encountered in edge devices while still supporting on-device gradient updates. Among the feasible candidates that satisfy the device’s resource constraints, we select the smallest model that achieves an accuracy within a small tolerance of the best-performing candidate. This strategy effectively balances efficiency and performance, ensuring that the chosen classifier maximizes on-device feasibility without incurring a significant loss in learning effectiveness.}

Formally speaking, the iterative process can be viewed as an optimization problem described as
\revs{
\begin{equation}
C=\argmin_{\{m_1,\dots,m_K\}} \; \mathbb{P}(m_k)\,\mathbb{B}~~\text{s.t.}~~\mathbb{P}(m_k)\,\mathbb{B} \le \mathbb{M}~\text{and}~\text{acc}(m_k) \ge \tau^{*},
\end{equation}
}
where $\mathbb{P}(m_k)$ provides the number of parameters of the candidate architecture $m_k$, $\mathbb{B}$ is the number of bytes used to store the parameters and intermediate results in computations, $\mathbb{M}$ is the memory available in the \iot device for FL participation, \revs{and $\tau^{*} = \max_{k}\text{acc}(m_k)$}.  The parameter $\mathbb{M}$ can be measured and logged as empirical observations of the \iot memory usage, to be sent to the server as we discussed in Step~\textcircled{\raisebox{-.9pt} {1}} in \figurename~\ref{fig:PbFL}.

\begin{algorithm}[]
\small
\SetAlgoLined
\setcounter{AlgoLine}{0}
\nl
\textbf{Inputs:}
        (1) $\mathcal{A}$: the set of $A$ clients each having two devices, one \iot and one \ap, (2) $D^{k}$:  local dataset for each client $a\in\{1, \dots, A\}$, (2) $f_E$: the encoder part with parameters $W_E$, (3) $f_C$: the classifier part with parameters $W_C$, (4) $R$: the total number of FL rounds.
     \vspace{0.1cm}\\
    \nl \For{$r \in \{1, \dots, \sfrac{R}{2}\}$ rounds}{
    \nl $K$ clients $\Leftarrow$ randomly select $K$ clients from a total of $A$, using uniform sampling
    \vspace{0.1cm}\\
    \emph{\textbf{---~Training on \iots~---}}
    \\
    \nl \For{$k \in \{1,...,K\}$}{
        \nl $\mathbf{X}^k, \mathbf{y}^k \leftarrow D^{k}$ \textrm{\em(all the samples in the local dataset of client $k$ \iot)} \\
        \nl $\ell^k = \mathcal{L}(f_E \circ f_C(\mathbf{X}^k), \mathbf{y}^k)$ \texttt{\em(forward pass to compute per-sample loss values)} \\
        \nl $D_N^{k}, D_C^{k}, D_M^{k} \Leftarrow$ based on $\ell^k$, perform data selection as detailed in Section~\ref{subsec:data_selection} \\
        \nl \For{all $(\mathbf{X}^k_j, \mathbf{y}^k_j) \in D_C^{k}$}{
            \nl $g^k_C = {\partial \ell^k}/{\partial W_C}$
             \texttt{\em(compute $f_C$'s gradients on $D_C^{k}$ to update $f_C$)}\\
             \nl $W_{C}^{k} = Optimizer(W_C^k, g^k_C)$ \texttt{\em (update $f_C$'s parameters using gradients)}
        \\
             \nl $D^+ \Leftarrow$ based on $g^k_C$, perform data selection as detailed in Section~\ref{subsec:data_selection} \\
        \nl $D_M^{k} = D_M^{k} \cup D^+$ \texttt{\em(enhance  $D_M^{k}$ by high-value gradients data points)}
        }
    }
    \vspace{0.1cm}
    \emph{\textbf{---~Server Side  Aggregation on the Classifier Part~---}} \\
    \nl $W_C \leftarrow \frac{1}{K} \sum_{k \in K} W_{C}^{k}$
    \\
    \nl $W_C^k \leftarrow W_C$ \texttt{\em(update all clients with new $W_C$)}
    \vspace{0.1cm}\\
    \emph{\textbf{---~Training on \aps~---}} \\
    \nl \For{$k \in \{1,\dots,K\}$}{
        \vspace{3pt}
        \nl $\mathbf{X}^k, \mathbf{y}^k \leftarrow D_M^{k}$ \texttt{\em(all selected samples for training $f_E$ on client $k$ \ap)}
        \\
        \nl $\ell^k = \mathcal{L}(f_E \circ f_C(\mathbf{X}^k), \mathbf{y}^k)$
        \texttt{\em(forward pass to compute per-sample loss values)}
        \\
        \nl $g^k = {\partial \ell^k}/{\partial (W_{E},W_{C})}$
             \texttt{\em(compute gradients on $D_M^{k}$ to update $f_E$ and $f_C$)}
             \\
        \nl $W_E^{k},W_C^{k} = Optimizer(W_E^k, W_C^{k}, g^k)$ \texttt{\em (update all parameters using gradients)}        
    }
    \vspace{0.1cm}
    \emph{\textbf{---~Server Side Aggregation on both Encoder and Classifier Parts~---}} \\
    \nl $\{W_E,W_C\} \leftarrow \frac{1}{K} \sum_{k \in K} \{W_E^k,W_C^k\}$
    \\
    \nl $\{W_E^k,W_C^k\} \leftarrow \{W_E,W_C\}$  \texttt{\em(update all clients with the new aggregated model)}
    }
    \vspace{0.1cm}
    \nl return $\{W_E,W_C\}$
\caption{\oursystem: Multidevice Federated Learning via Partition-based Training and Data Selection}
\label{alg:training_aggr}
\end{algorithm}

\subsection{\textbf{Data Selection}}
\label{subsec:data_selection}

We determine the importance of each data sample, along the forward pass, to decide whether to further perform (the more costly) backward pass on this sample or not. 
Such data selection satisfies design requirements by improving both storage efficiency and training efficiency.
We use a combination of (i) the {\em loss value} and (ii) the {\em norm of last-layer gradients} to measure the sample's importance.

\subsubsection{Loss-based Selection.} All available data $D$ at round $r$ are fed into the model $M$ to compute their loss values $\ell_r(f_M(\mathbf{X}),\mathbf{y})$. 
We consider lower loss values as an indication of being less important, and higher loss values show the importance of the data~\cite{jiang2019accelerating}.
To this end, at round $r$ for the current $\ell_r$, we drive a cumulative distribution function~(CDF$^{\ell}_{r}$). Then at round $r+1$, the probability of discarding data, $\mathcal{P}_N$, and the probability of feeding data to train the complete model $M$ on \ap, $\mathcal{P}_M$, are defined as: 
\begin{equation}\label{eq:data_selection_loss}
    \begin{cases}
    \mathcal{P}_N(\ell_{r+1}) = 1-[\text{CDF}^{\ell}_{r}(\ell_{r+1})]^\alpha \\
     \mathcal{P}_M(\ell_{r+1}) = [\text{CDF}^{\ell}_{r}(\ell_{r+1})]^\beta
    \end{cases}
\end{equation}
where $\alpha$ and $\beta$ parameters determine the level of selectivity. At round $r+1$, for a sample with loss value $\ell_{r+1}$, the sample is selected into $D_N$ with probability of $\mathcal{P}_N(\ell_{r+1})$.  Thus, the lower the loss value for a sample, the greater the probability for being discarded. Similarly, $\mathcal{P}_M$ selects samples with the highest losses as $D_M$; which means that the higher the loss value for a sample, the greater the probability of being chosen to train the entire model. Samples that are not selected for $D_N$ or $D_M$ are added to $D_C$ to locally train $C$ on the \iot. We use a fixed-size queue for the CDF$^{\ell}_{r}$ to dynamically keep track of loss values so that the $\mathcal{P}_N$ and $\mathcal{P}_M$ for the current sample can be efficiently computed using its loss and to save computational resources similar to prior work~\cite{oort}. 

\subsubsection{Gradient-based Selection.} 
For samples that are \emph{not} selected through $\mathcal{P}_N$ and $\mathcal{P}_M$, and to avoid consuming extra resources of \iots, we can further compute the gradients of the classifier's last layer. The norm of these gradients gives us a useful hint about the sample's importance while requiring much less computation than computing all layers' gradients~\cite{katharopoulos2018not}.
The last layer's norm at round $r$ is produced during training of the classifier by $g_r = \frac{\partial \ell_r}{\partial W_r}$, where $W_r$ is the weights of the classifier $C$'s last layer at round $r$.
Following the same idea, we derive a CDF$^g$ for gradient norm values and build a queue to keep track of the computed norm of gradient values. Samples with larger gradients have larger impacts on the model's weights; thus, we also keep these samples for training the entire model $M$ as $E$ might learn new ``features'' from them.
Thus, the probability of adding a sample to $D_M$ at round $r+1$ is defined as:
\begin{equation}\label{eq:data_selection_norm}
      \mathcal{P}_{M}^{+}(\|g_{r+1}\|) = [\text{CDF}^{g}_{r}(\|g_{r+1}\|)]^\gamma
\end{equation}
where $\gamma$ is used for customizing the selection rate. Samples with high norm are selected with $\mathcal{P}_{M}^{+}(\|g_{r+1}\|)$ and then added to data $D_M$.  Our dynamic strategy, in combining loss and last-layer gradient norm, enables \oursystem to achieve a better trade-off between computation cost and selection performance on \iots. Moreover, one may substitute this module with a different data selection technique that could also be dependent upon the specific use case.

\revs{While our proposed data selection strategy is primarily empirical, it is motivated by the theoretical principles of importance sampling in deep learning. Prior work has shown that sampling data points in proportion to their loss values or gradient magnitudes can approximate the optimal sampling distribution that minimizes the variance of stochastic gradient estimates and thus accelerates convergence~\cite{katharopoulos2018not, jiang2019accelerating}. In particular, high-loss samples tend to carry larger gradient norms and contribute more substantially to the update direction, while samples with small losses or small gradient norms contribute less. Our combination of loss value and last-layer gradient norm therefore aligns with these theoretical insights: it adaptively emphasizes samples expected to produce high-informative gradients, balancing computational cost with learning efficacy. Nonetheless, we acknowledge that our current formulation does not provide a formal convergence guarantee. Developing theoretical guarantees through variance-reduction or convergence-rate analysis is a promising direction for future work.}

\subsection{\textbf{Partition-based Training and Aggregation}}
\label{subsec:fl_training}
\revs{The results of \S\ref{sec_motiviating_rpi} highlight why we freeze the encoder and only train the classifier on \iots. Training just the classifier consumed far less memory and CPU, comparable to running an inference pass, and ran much faster than full-model training. Since most \iots have far less memory than even a RaspberryPi (e.g. a few MB of storage or only a few hundred KB of SRAM), full-model training on \iots is generally infeasible. By freezing the encoder on \iots, \oursystem aims to reduce the on-device computation and memory requirements, allowing each \iot to train only the compact classifier locally. The more powerful \ap (smartphone or router) can then periodically update the encoder on the offloaded samples, so that the overall model can still learn richer features even though each \iot only fine-tunes the classifier part.}

Therefore, our training is conducted on both \iots and \aps. While \iots only train the classifier $C$ on data $D_C$, \aps train the complete model $M$ on data $D_M$.
Such a procedure allows the model to be (partially) updated when \iots are offline, in addition to the full updates when \iots have a connection to the internet. 
This better utilizes \iots' spatiotemporal richness and consequently improves the training efficiency in our design requirements.

Algorithm~\ref{alg:training_aggr} shows partition-based training and aggregation in \oursystem, running $\sfrac{R}{2}$ rounds, considering that \iot training and \ap training proceed iteratively in one round. All training is conducted on the client's devices either on \iots or \aps. Then with all updated weights, the server performs Federated Averaging (FedAvg)~\cite{floriginal} to obtain one aggregated global model. The new global model needs to be distributed to all clients \iots and \aps such that (i) the training on \ap has an updated classifier, and (ii) the training on \iot in the next round has both an updated encoder and classifier.

\revs{\oursystem is the first FL framework to simultaneously combine model partitioning, data selection, and on-device training. Our approach of partitioning the DNN into encoder and classifier, selecting a subset of data on-device, and performing lightweight on-device updates, has not been addressed jointly in earlier work. In contrast, existing hierarchical FL schemes either train full models on each device or only aggregate updates at the AP without splitting the model.}

\section{Experimental Setup}\label{sec_exp_setup}

{\bf Datasets.}
We use six commonly used datasets. CIFAR10 and CIFAR100~\cite{krizhevsky2009learning} that both include 50K training samples and 10K test samples, with 10 and 100 classes, respectively. EMNIST~\cite{cohen2017emnist} that includes 112K training samples with 47 classes. UCIHAR~\cite{Anguita2013APD} is a widely used dataset of 30 users performing 6 daily activities; data from the accelerometer and
gyroscope sensors were collected by a smartphone worn on the waist. Data from 21 users is used for training, and that of the other 9 users for testing purposes. MotionSense~\cite{malekzadeh2019mobile} also includes accelerometer and gyroscope data from 24 users with a smartphone in the pocket of the trousers who performed 6 activities in 15 trials. We use as test data one trial session for each user and as training data the remaining trial sessions (\eg one trial of ``walking'' of each user is used as test data and the other two trials are used as training). PAMAP2~\cite{reiss2012introducing} dataset contains data of 13 different physical activities, performed by 9 subjects wearing 3 devices and a heart rate monitor. We use the training-test split provided by the PersonalizedFL library~\cite{PersonalizedFL}.

{\bf Models.} For image classification datasets, we select four DNNs commonly used in mobile/edge-oriented literature: (1)~EfficientNet-v2~\cite{tan2021efficientnetv2}, (2)~MobileNet-v3~\cite{howard2019searching}, (3)~MNASNet~\cite{tan2019mnasnet}, and (4)~ShuffleNet~\cite{ma2018shufflenet}, with 5.3M, 2.2M, 2.5M, and 1.4M number of parameters respectively. For human-activity recognition datasets, we borrow the ConvNet architecture proposed in~\cite{chang2020systematic}. All these models consist of an encoder made of multi-layer CNNs followed by a classifier.

{\bf FL Settings.} 
We use Flower~\cite{beutel2020flower}, a customizable open-source FL framework (Python~v3.7, Ray~v1.11, and Torch~v1.12). We run all simulations on a server with 80 Intel Xeon(R) E5-2698 CPUs, 8 Tesla V100 GPUs (16GB), and 504GB system RAM. For all simulations, we use 100 communication rounds, where at each round, \iots and \aps successively train the model. 
Unless specified otherwise, we consider 100 clients (each having one \ap and one \iot).
We randomly sample 10\% of the clients for training in each round in our initial experiments and later show that our results hold when the client participation ratio is gradually increased to 100\%.
Data ($D$) are partitioned using Latent Dirichlet Allocation (LDA)~\cite{blei2003latent, beutel2022flower} without resampling (LDA-Alpha=1000). We refer to prior research~\cite{blei2003latent} about details of LDA's generative process. After such partition, each client owns $\sfrac{|D|}{100}$ local samples.

For each client, we randomly select half of the samples and use them for local training of the model.
For both \aps and \iots, we consider by default the number of epochs to be 3 with a batch size of 64.
The other half of the samples are available with more offline time to simulate the further collection of extra data.
\aps and \iots also train more epochs on the extra data while offline. In this setup, we define every unit offline time contributing to one more epoch of training on $20\%$ of these extra local samples.

\begin{figure}[t]
    \centering
    \includegraphics[width=\columnwidth]{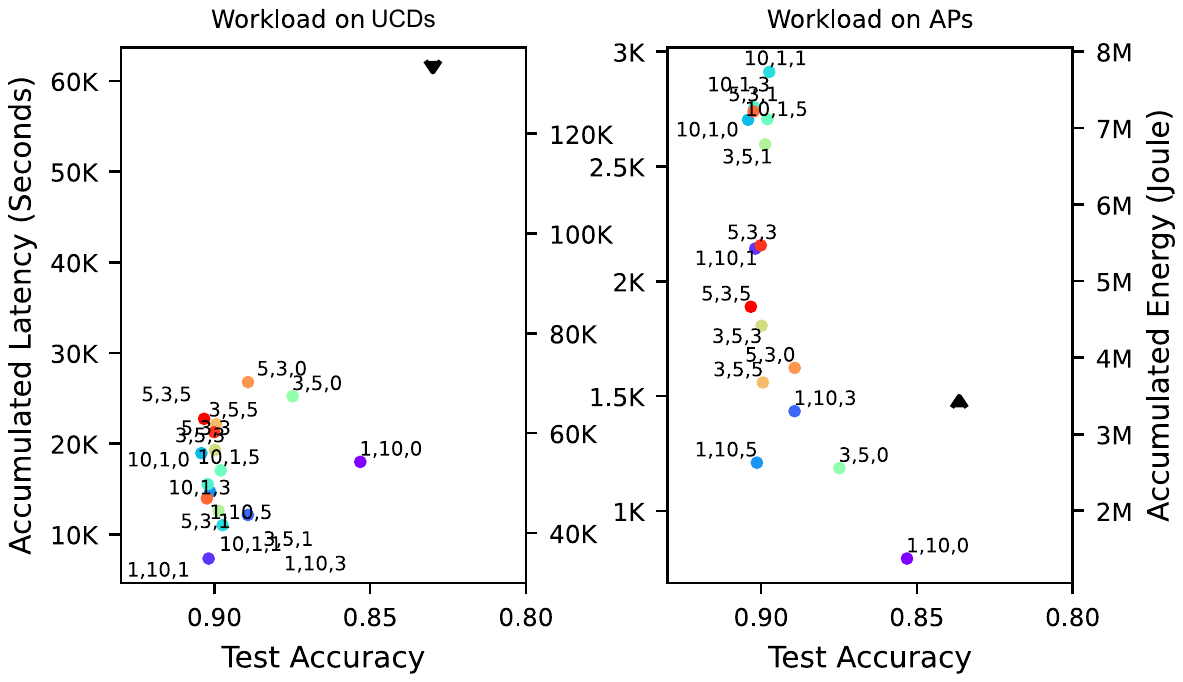}
    \caption{   Test accuracy when using different values for parameters $\alpha$, $\beta$, $\gamma$ in data selection. The $\blacktriangledown$ point in the left figure is the \iot training, and the $\blacktriangle$ point in the right figure is the \ap training. It is found that with data selection, \oursystem can always achieve higher accuracy than both \iot training and \ap training.}
    \label{fig:data_selection_args}
\end{figure}

\revs{To set the hyperparameters for Eqs~\eqref{eq:data_selection_loss}-\eqref{eq:data_selection_norm}, we run an experiment on CIFAR-10 with MobileNet-V3. We evaluate the trade-off between test accuracy and the workload imposed by data selection on both \iots and \aps. This experiment can be executed by the server as a preparatory step before deploying \oursystem, using small representative datasets that approximate the clients’ data distribution. The results in \figurename~\ref{fig:data_selection_args} show that performing data selection reduces latency and energy consumption on \iots by over 80\%. Although this approach increases the workload on \aps, this is acceptable given their higher computational capacity, since we observe a substantial improvement in model accuracy. In our experiments, we set hyperparameter $\alpha,\beta,\gamma$ as $5,3,0$, respectively, as this configuration represents the fair balance between accuracy and resource efficiency.}

{\bf Baselines.} \textbf{(1) \ap Training.} We consider \iots as data collection apparatuses only and therefore, \iots upload collected and stored data samples to their connected \aps, so that the training only happens on \aps. In this case, the complete model (both the encoder and classifier) is updated because APs in general are considered to have sufficient resources.
\textbf{(2)~\iot Training.} \iots do not upload the collected data samples. Instead, they conduct training on the data locally. However, as \iots are typically resource-constrained, only the classifier is considered to be trained on-device. This means that the encoder part is considered to be frozen in this setting.

{\bf Mobility Model.} 
In our case, the idea of the mobility model signifies how the mobility patterns of the \iots (or users) impact the connectivity of the \iot to its \ap and eventually to the internet. We define an exclusive Online Association Matrix $\Omega$, which is a binary matrix representing the user's exclusive location at a given time instance. Mathematically, $\Omega$ is a binary matrix $\mathcal{T}\times\mathcal{S}$ where $\mathcal{T}$ and $\mathcal{S}$ represent the temporal and spatial granularity of $\Omega$ marix, respectively. For example, the rows may represent the time zones of the day (early morning, morning, afternoon, evening, and late night), during which the user moves between three locations like home, office, and a public park, represented by the columns of $\Omega$. Furthermore, we ensure that each row's sum equals unity, which means the user is present exclusively at a unique location in a given temporal instance. Finally, to simulate the connectivity patterns, we define a connectivity matrix $\lambda$, a single row-vector of dimension $\mathcal{S}$, denoting the connectivity probability across different locations.
More specifically, for evaluating \oursystem during mobility, we generate the global connectivity matrix $\lambda$ and the generate matrix $\Omega$, both chosen uniformly at random, for each user to simulate the mobility scenario.

\begin{figure}[t!]
    \centering
    \includegraphics[width=\columnwidth]{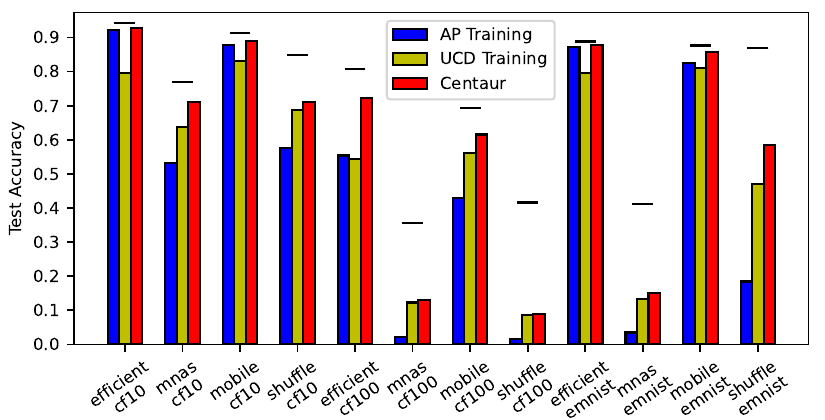}
    \caption{   Accuracy of the best classifier for four different encoders, trained on CIFAR10, CIFAR100, and EMNIST. Dash lines (\blackline) above depict the upper-bound accuracy when \iot devices have no resource and connectivity constraints.}
    \label{fig:all_accuracy}
\end{figure}

\section{Evaluation Results}
\label{sec:eval_results}
We present the results when running \oursystem compared to conventional FL training methods and compare the approaches using different metrics. 
We also analyze the impact of data/participation heterogeneity and spatiotemporal coverage on training methods.

%
\subsection{Metrics} \label{setup}
In addition to the accuracy of the trained global model on a held-out test set we evaluate the performance of our FL framework, compared to other FL alternatives, using the following metrics.

{\bf 1)} \textbf{Accuracy.}
This is the classification accuracy of the trained model on a test set hosted by the central server. For a model obtained at the end of each training round, the test accuracy is computed as 
$$acc \equiv \frac{1}{T}\sum_{i=1}^{T} \mathbb{1}\big(\argmax\big(f_E \circ f_C(\mathbf{X}_i)\big) = \mathbf{y}_i \big),$$
where test set has $T$ pairs of $(\mathbf{X}_i, \mathbf{y}_i)$ and $\mathbb{1}(\mathrm{C})$ denote the indicator function that outputs 1 if condition $\mathrm{C}$ holds.

{\bf 2)} \textbf{Multiply–Accumulate~(\MAC)}.
This operation is a common step that computes the product of two numbers and adds that product to an accumulator (\mbox{$a \leftarrow a + (b\times c)$});
a fundamental operation for both any DNN layers during training and inference. We use \texttt{fvcore}~\cite{fvcore} library to compute the number of MAC operations. Since \texttt{fvcore} only supports counting MACs in a forward pass, and to count MACs in a backward pass, we use the heuristic that FLOPs (\ie~double MACs) ratio of the backward-forward pass is typically between 1$\times$ and 3$\times$ and most often is 2$\times$ based on models' specific layer types, according to previous observations~\cite{marius2021backprop}.

\begin{figure}[t!]
    \centering
    \begin{subfigure}{\columnwidth}
    \includegraphics[width=\columnwidth]{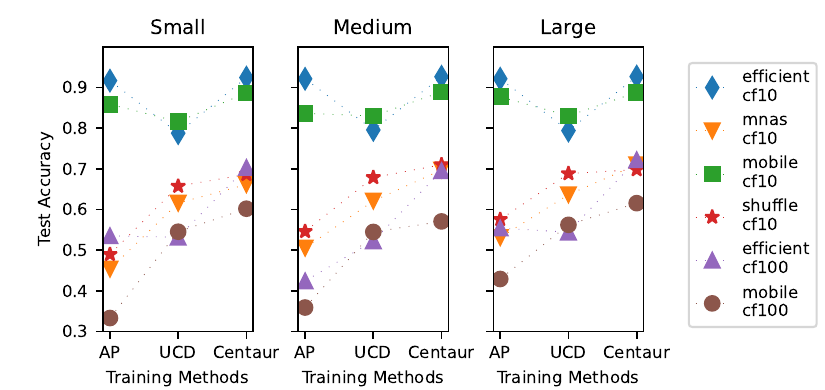}
    \end{subfigure}
    \begin{subfigure}{\columnwidth}
    \centering
    \includegraphics[width=\columnwidth]{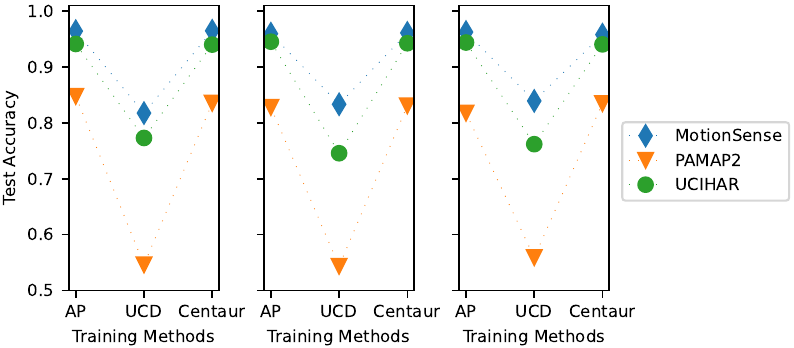}
    \end{subfigure}    
    \caption{   \oursystem performance vs. other baselines for different sizes of classifiers: small, medium, and large. (Top) image classification, and (bottom) HAR datasets.
    }
    \label{fig:classifier_acc}
\end{figure}

{\bf 3)} \textbf{\Bandwidth}.
As the model size (\ie~the encoder or the classifier) can be different in each FL round, we use \texttt{fvcore} to count the number of parameters that are communicated in each round. Based on the model size and number of communications among \iots, \aps, and the server, we compute the amount of bandwidth that is consumed. Also, we count the number of sample points that are uploaded from \iots to \aps.

\begin{figure*}[t!]
    \centering
    \begin{subfigure}{\textwidth}
    \centering
    \includegraphics[width=\textwidth]{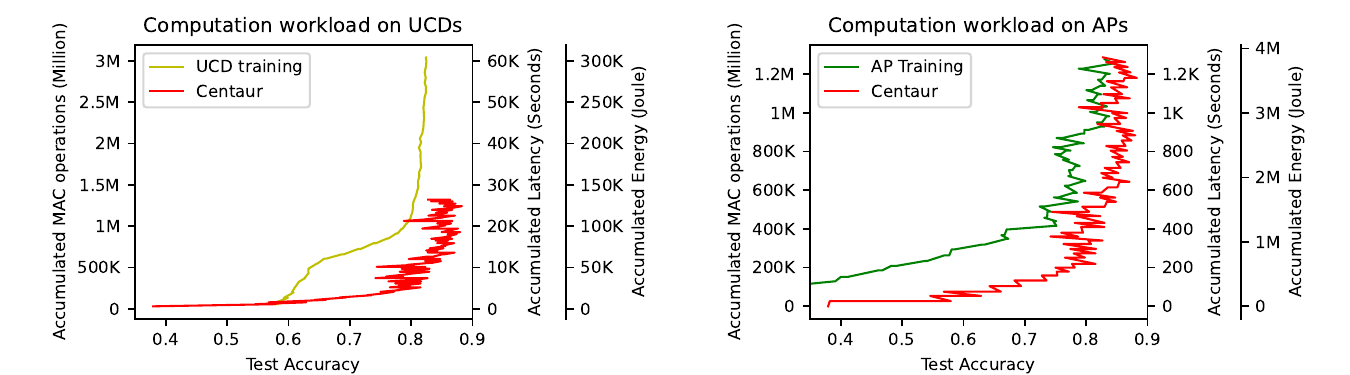}
    \end{subfigure}
    \begin{subfigure}{\textwidth}
    \centering
    \includegraphics[width=\textwidth]{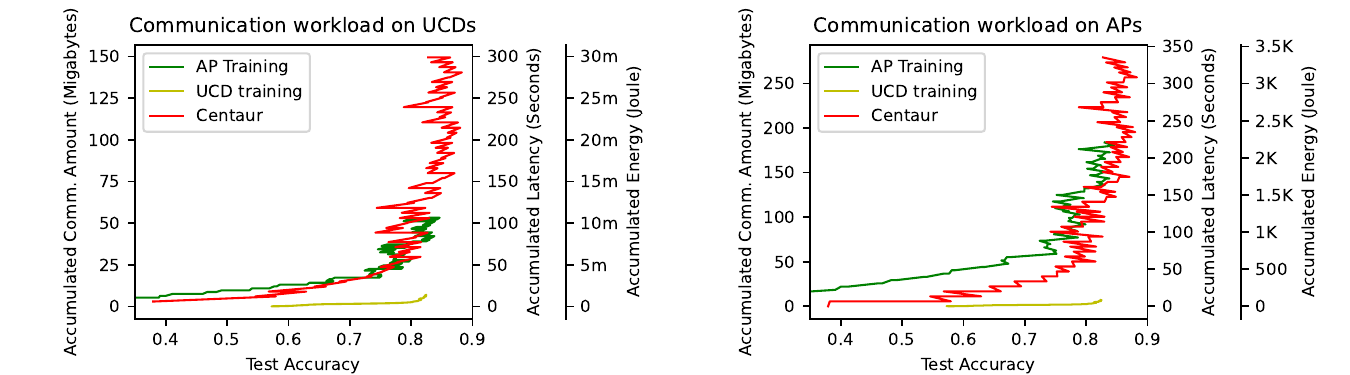}
    \end{subfigure}
    \caption{Computation workload of model training on usage-constrained devices (Top Left) and access points (Top Right), and Communication workload of model/data sharing of usage-constrained devices (Bottom Left) and access points (Bottom Right) when achieving specific test accuracy, reported on MobileNetv3 and CIFAR10.}
    \label{fig:mac_latency_accuracy}
\end{figure*}

{\bf 4)} \textbf{\Latency}.
The latency usually has a linear relation with MAC operations, due to the lack of specialized accelerators~\cite{liberis2021munas,lin2022device}. Thus, we estimate the latency of model training based on the processor's frequency as - 
%
$\text{Latency} = c * \frac{\text{Total MACs}}{\text{Processor Frequency}}$.
The ratio between MAC operations and the processor's instructions, $c$, is typically between $1$ and $2$ based on specific instruction sets/compilers. For simplicity, we assume that each MAC operation translates to two instructions in an MCU~(\ie~$c=2$).
Notice that existing processors (\eg~Intel's Load Effective Address) complete one MAC in one instruction~\cite{x86assembly}), and this only scales the experimental results and does not change the conclusions.
Finally, communication latency is estimated based on the total amount of data needed to be transmitted, divided by the up-link speed or the down-link speed of devices.

{\bf 5)} \textbf{\Energy}.
This is the total execution time multiplied by the processor's consumed power per unit time. The energy consumption of communication can also be calculated based on the total time of transmitting data multiplied by the transmitters' power per unit time. We set the values for \iots and \aps in our simulations as follows. {We assume a typical \ap to have a CPU frequency of 2GHz, storage capacity of 4GB, power capacity of 1.5mW/MHz with an uplink speed of 10Mbit/s, downlink speed 100Mbit/s, and communication energy of 10W. For a typical \iots, we assume a CPU frequency of 100MHz, storage capacity of 5MB, power capacity of 0.05mW/MHz with an uplink speed of 2Mbit/s, downlink speed of 2Mbit/s, and communication energy of 0.0001W.}

\subsection{Model Accuracy}
We use the CNN backbone of the four benchmark models described in \S\ref{sec_exp_setup} as the encoders. 
We examine three classifiers: {\em small} that is only one fully connected~(FC) layer of size $z$ (number of classes), {\em medium} that has two FC layers of size 64 and $z$, and {\em large} that has two FC layers of size 128 and $z$.

We permute the four encoders and three classifiers and report the test accuracy of \oursystem compared to the two other baselines: \ap training and \iot training.
In \figurename~\ref{fig:all_accuracy} we report the highest test accuracy for each encoder across all three classifiers for three datasets.
Results show that \oursystem outperforms \ap training by $0.53\% \sim 40.15\%$ and \iot training by $0.45\% \sim 17.78\%$, depending on the settings.
There are several settings in which all FL training methods cannot reach a good accuracy, \eg~NASNet and ShuffleNet on CIFAR100, probably because of their relatively small model sizes compared to data complexity. \oursystem still achieves better performance compared to the baselines.

In \figurename~\ref{fig:classifier_acc} we present the accuracy of different classifiers for each encoder. 
In the top plot, the results of training on CIFAR10 and CIFAR100 with an accuracy higher than $30\%$ are presented for better visibility. In the bottom plot, we report the results of UCIHAR, MotionSense, and PAMAP2. The results show that \oursystem outperforms both \iot training in all classifiers' sizes.
The test accuracy also tends to be similar across small, medium, and large classifiers. 
In addition, the classifier's sizes may have less impact on more sophisticated encoders (\eg~EfficientNet and MobileNet), which is also observed in a previous work~\cite{mo2021ppfl}. This may be because an appropriate encoder already produces high-quality features for unseen data that are easy to learn (\eg~CIFAR10), and in such a case, classifier sizes do not make any difference in test accuracy. Similar patterns are observed in HAR datasets. Notice that for HAR datasets, there is currently a lack of publicly accessible pre-trained encoders to start FL with. As a result, we can observe a more significant performance gap compared to image classification, particularly for more complex datasets, \eg PAMAP2. This lack of pre-trained encoders can also be the reason that, in HAR datasets, the performance of \oursystem and \ap are almost the same. 

\revs{These results emphasize that with a properly pre-trained backbone, continual re-training of the encoder on \aps is neither necessary nor desirable. In transfer learning, it is standard to fix, or only occasionally fine-tune, the encoder and update the lightweight classifier head to adapt to local data. In \oursystem, we follow this principle: \iots perform frequent classifier-only updates to rapidly personalize to local data, while \aps periodically update the encoder to correct drift in feature extraction. This approach achieves competitive accuracy with lower communication and computation costs: frequent classifier updates capture personalization and mobility, while periodic encoder updates maintain convergence and generalization. Future work can explore how varying update frequencies affect convergence trade-offs.}

\begin{figure}[t!]
    \centering
    \includegraphics[width=\columnwidth]{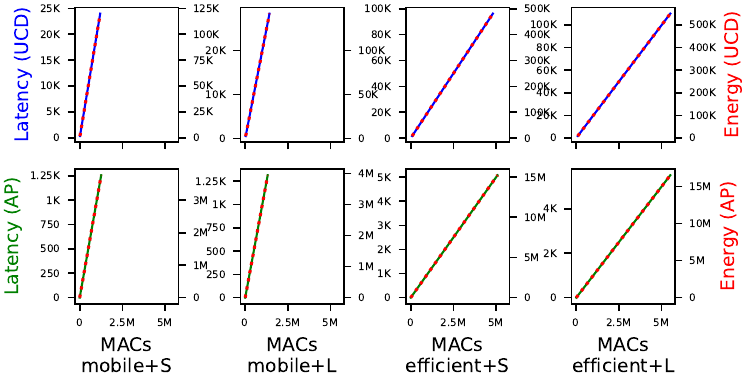}
    \caption{Cost linkage analysis among MAC operations, latency, and energy consumption.}
    \label{fig:cost_linkage}
\end{figure}



\subsection{Efficiency}


{\bf Cost-Accuracy Trade-off.}
Here, with CIFAR10, MobileNet-v3 as the encoder, and a medium size for the classifier, we compute the test accuracy for a range of \MAC, \Bandwidth, \Latency, and \Energy budgets. The average size of each sample of CIFAR10 is $30KB$. \figurename~\ref{fig:mac_latency_accuracy} (top) shows on both \iot and \ap and across accuracy levels, \oursystem achieves lower \MAC and \Latency than both \iot training and \ap training.  \figurename~\ref{fig:mac_latency_accuracy} (bottom) shows that communication \Bandwidth and \Latency of \oursystem is almost the same as \ap training; however, \oursystem can achieve a higher test accuracy along with more training steps when consuming more resources. \revs{\oursystem uses much more communication resources than \iot training, as \oursystem transmits both the classifier, although it is much lighter than the encoder, and data samples}. However, the accuracy of \iot training cannot go further than $83.10\%$, while \oursystem can reach up to $89.90\%$ accuracy.  \revs{More reduction in uplink data transmission of \iots can be achieved by a more advanced data selection strategy, lowering the communication burden on constrained devices. This also leads to a notable decrease in overall communication cost compared to \ap training, as less data is exchanged between \iots and \aps while improving model accuracy.} We remark that model training consumes much more energy and causes much more latency than communication.

\begin{figure}[t!]
    \centering
\includegraphics[width=\columnwidth]{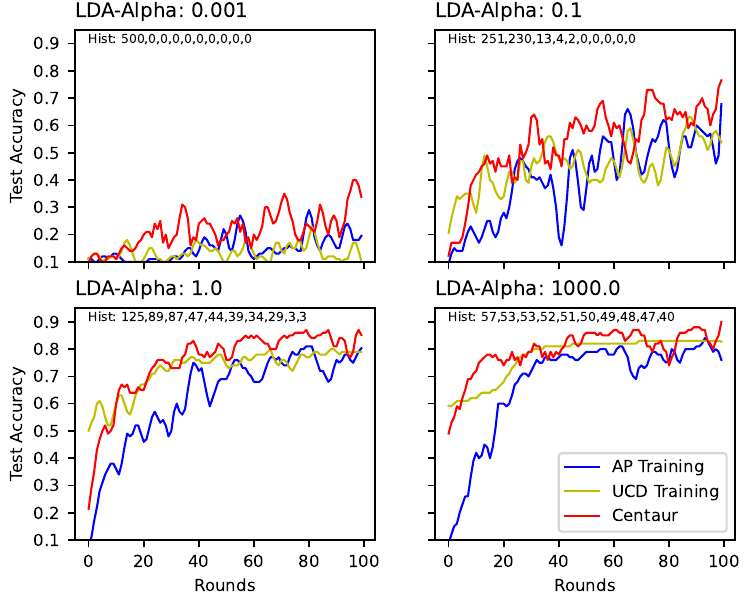}    
    \caption{The impacts of imbalanced data partitioning on test accuracy of \ap training, \iot training, and \oursystem, on MobileNetv3 with CIFAR10 (annotated texts are examples of the classes present on a \iot)}
    \label{fig:unblanced_1}
%
\includegraphics[width=\columnwidth]
    {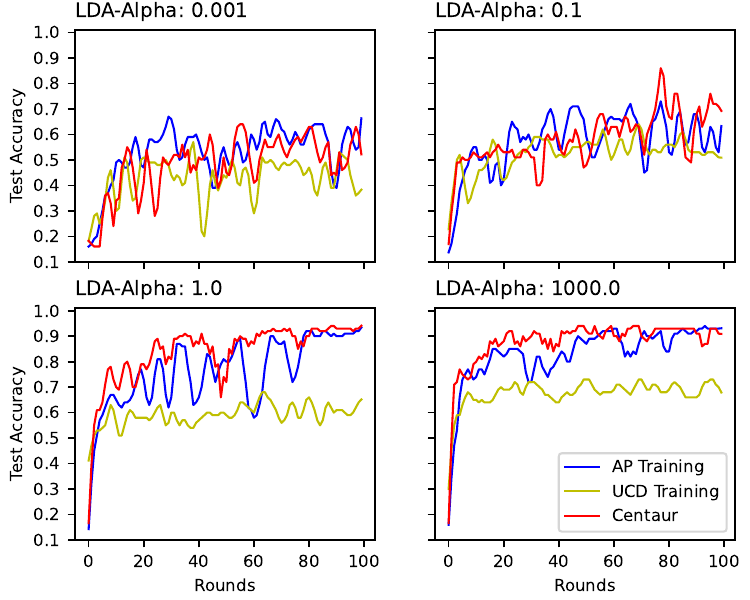}
    \caption{The impacts of imbalanced data partitioning on test accuracy of \ap training, \iot training, and \oursystem, on ConvNet and UCIHAR dataset dataset.}
    \label{fig:unblanced_2}
\end{figure}

{\bf Correlation among Cost Metrics.} 
To show the connection between cost metrics, we plot \MAC, \Latency, and \Energy in \figurename~\ref{fig:cost_linkage}. We observe that both \Latency and \Energy have linear relationships with \MAC. Besides, \Latency is also linearly correlated with \Energy, in such a way that these two entirely overlap when we appropriately scale the y-axes. 
We remark that \Bandwidth also has a similar correlation. Such linear relationships are because of the assumption we made when computing \Latency and \Energy; however, the actual cost and the relationship among these metrics might not deviate much in practice. Based on such correlations, we can add \Latency (or \Energy) of model training as well as the communication \Bandwidth to get the overall workload. 
The results shown in \figurename~\ref{fig:intro_sum} indicate that the workload still follows a similar pattern as the computation in model training, because the training workload significantly overweight the communication workload.


\subsection{Data and Participation Heterogeneity}
{\bf \oursystem is robust to data heterogeneity.}
To create imbalanced non-IID data partitions among clients, we use LDA as defined in \S~\ref{setup}. Next, we set different values for LDA-Alpha to manipulate the levels of non-IID data partitions. 
\figurename~\ref{fig:unblanced_1} and \figurename~\ref{fig:unblanced_2} show the results when LDA-Alpha are changing from $0.001$, to $0.1$, to $1$, and to $1000$. The smaller LDA-Alpha is, the less balanced the dataset is. We also annotate some examples of class distribution in \figurename~\ref{fig:unblanced_1}, demonstrating that LDA-Alpha$=0.001$ generates almost $1$ class per client, while LDA-Alpha$=1000$ generates almost uniform classes per client.
Results show that in all cases \oursystem can reach higher test accuracy than both \ap and \iot training.
We remark that due to the retraining of the encoder, both \oursystem and \ap training can cause more fluctuations than \iot training; this is obvious when data distribution tends to be IID (\eg~when LDA-Alpha$=1000$ in \figurename~\ref{fig:unblanced_1}).

\begin{figure}[t!]
    \centering
    \includegraphics[width=\columnwidth]{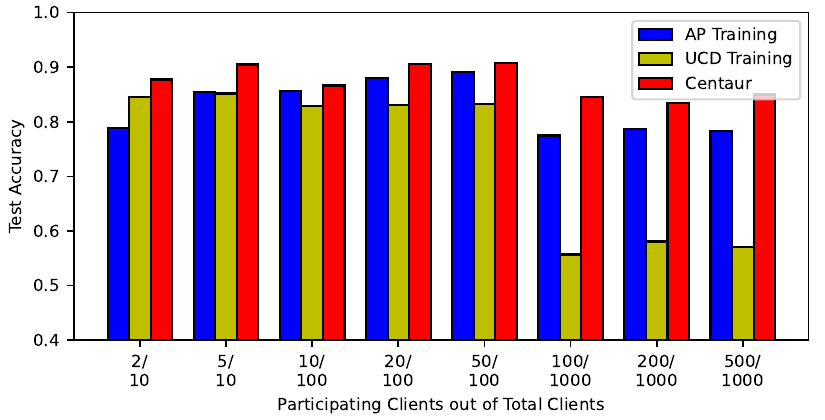}
    \caption{The impact of participating vs. total clients on test accuracy for MobileNetv3 with CIFAR10.}
    \label{fig:clients_num}
\end{figure}

{\bf \oursystem scales well with higher total and participating clients.}
Here, we scale the FL training problem with the number of total clients from 10 to 1000. $10\%$, $20\%$, and $50\%$ of them are selected respectively as participating clients in each round. \figurename~\ref{fig:clients_num} shows \oursystem achieves higher test accuracy compared to both \iot training and \ap training.
Specifically, with a higher participation rate, all training methods tend to have better performance. However, with a much larger number of total clients (\eg~1000), the test accuracy is reduced, because the same size of data is partitioned into more portions. 
Also, with such more fragmented data partitioning, \oursystem has much accuracy gain when compared to \iot training. \revs{This experiment focuses on test accuracy and does not account for the orchestration costs of running \oursystem in dynamic mobile environments. However, such scalability challenges are not unique to \oursystem but are inherent to any FL framework. We also note that practical deployment at larger scales necessitates additional server-side  coordination. \oursystem is designed to be modular, enabling integration into existing FL stacks. Nevertheless, evaluating the end-to-end overhead of managing varying participation rates and asynchronous aggregation remains open.}

\subsection{Performance of \oursystem under Mobility}

\begin{figure}[t!]
    \centering
    \includegraphics[width=\columnwidth]{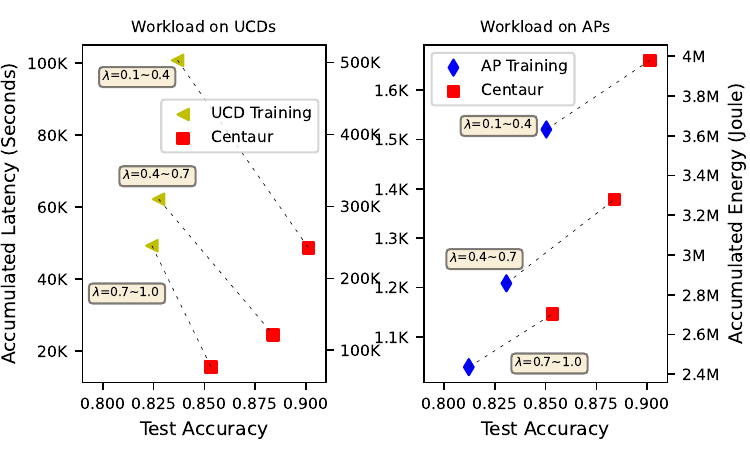}
    \caption{The performance of access point (AP) training, usage-constrained device (\iot) training, and partition-based federated learning (\oursystem) under different connection probabilities $\lambda$, reported on MobileNetv3 and CIFAR10.}
    \label{fig:mobility}
\end{figure}
We study when the connection probability $\lambda$ changes based on the mobility model defined in \S\ref{setup}. For a more readable visualization, we bucket $\lambda$ in three different ranges of $\lambda \in$ [0.1,0.4], $\lambda \in$ [0.4,0.7], and $\lambda \in$ [0.7,1.0] and report the corresponding accuracy when the \iots are mobile.
\figurename~\ref{fig:mobility} shows \oursystem always has higher efficiency than standard \iot training while incurring a higher latency than direct \ap training. Specifically, \oursystem gains $6.45\%$, $5.64\%$, and $4.11\%$ higher accuracy with $51.73\%$, $60.48\%$, and $68.35\%$ lower cost both in terms of energy and latency across the different ranges of $\lambda$. Interestingly, we also observe a rise in accuracy with lower connectivity probability. The main reason is the design of our experimental setup: we realistically assume that with limited connectivity, \iots devices can collect more data and perform data selection locally. Using this selected data, each device then trains the classifier for more epochs, which further improves its accuracy.

\revs{\section{Discussion}}
\revs{While \oursystem shows strong empirical performance and feasibility on constrained devices, some of our assumptions need further study. We outline key limitations, and opportunities for future work, below.}

\revs{(1) Centaur assumes each \iot has a trusted, resourceful personal \ap (\eg a smartphone or router) that is always available. This simplifies privacy and connectivity but may not hold in transient or multi-user settings.}

\revs{(2) We used RaspberryPi as a reproducible proxy for constrained devices, though it is more capable than many real \iots. Results may thus overestimate feasibility; testing on truly constrained hardware is needed to validate our claims.}

\revs{(3) Freezing the encoder and training only the classifier reduces on-device compute and memory but limits adaptability: if local data diverge from the pretrained one, classifier-only updates may be insufficient. Studying trade-offs in encoder vs. classifier update frequency  remains open.}

\revs{(4) Our on-device selection uses loss and gradient norm for importance sampling, but this remains heuristic without formal convergence or variance guarantees. It may also be sensitive to outliers, mislabeled, or out-of-distribution data, potentially biasing updates.}

\revs{(5) \oursystem reduces on-device workload but increases \ap computation and network use. We did not quantify orchestration and scaling costs with many asynchronous users. Although fewer samples are sent than baselines, uplink and energy costs for constrained devices and overall communication efficiency need further analysis.}

\revs{(6) We use FedAvg in our implementation, but \oursystem can accommodate other FL aggregation strategies. Layer-wise FL methods that perform model training in a partitioned manner, such as FedMA~\cite{wang2020federated}, can also be integrated. Also, we evaluated \oursystem on supervised tasks, yet a complementary approach can be suggested for unsupervised tasks.}

\section{Conclusions}
We propose an efficient training method that leverages data selection to improve learning and reduce computational costs in multidevice federated learning. Our approach allows partition-based training and aggregation on resource-constrained devices, supported by resourceful companion devices. Our evaluations over various scenarios confirm that our proposed solution improves model accuracy while reducing system overhead.


\bibliographystyle{ACM-Reference-Format}
\bibliography{main}

\end{document}